\documentclass[letterpaper, 10 pt, conference]{ieeeconf}  

\IEEEoverridecommandlockouts                              

\usepackage{graphicx} 

\usepackage{xcolor}
\usepackage{balance}
\usepackage{algorithm,algorithmic}

\usepackage{fancyhdr}

\fancypagestyle{FirstPage}{
\lfoot{\copyright 2023 IEEE. Personal use of this material is permitted. Permission from IEEE must be
obtained for all other uses, in any current or future media, including
reprinting/republishing this material for advertising or promotional purposes, creating new
collective works, for resale or redistribution to servers or lists, or reuse of any copyrighted
component of this work in other works.} 
}

\title{\LARGE \bf
    Introducing SSBD+ Dataset with a Convolutional Pipeline for detecting Self-Stimulatory Behaviours in Children using raw videos
}

\author{Vaibhavi Lokegaonkar$^{1}$, Vijay Jaisankar$^{1}$, Pon Deepika$^{1}$, Madhav Rao$^{1}$, T K Srikanth$^{1}$, \\ Sarbani Mallick$^{2}$, Manjit Sodhi$^{3}$
\thanks{$^{1}$Surgical and Assistive Robotics Lab, IIIT-Bangalore, Bangalore-560100, India. (e-mail: mr@iiitb.ac.in),
$^{2}$ Bubbles Centre for Autism, Bangalore,
$^{3}$ IBM India Software Labs
}%
}

\begin{document}

\maketitle
\thispagestyle{empty}
\pagestyle{empty}

\begin{abstract}
Conventionally, evaluation for the diagnosis of Autism spectrum disorder is done by a trained specialist through questionnaire-based formal assessments and by observation of behavioral cues under various settings to capture the early warning signs of autism. These evaluation techniques are highly subjective and their accuracy relies on the experience of the specialist. In this regard, machine learning-based methods for automated capturing of early signs of autism from the recorded videos of the children is a promising alternative.
In this paper, the authors propose a novel pipelined deep learning architecture to detect certain self-stimulatory behaviors that help in the diagnosis of autism spectrum disorder (ASD). The authors also supplement their tool with an augmented version of the Self Stimulatory Behavior Dataset (SSBD) and also propose a new
label in SSBD Action detection: \textit{no-class}. 
The deep learning model with the new dataset is made freely available for easy adoption to the researchers and 
developers community. An overall accuracy of around 81\% was achieved from the proposed pipeline model that is targeted for real-time and hands-free automated diagnosis. All of the source code, data, licenses of use, and other relevant material is made freely available in~\textcolor{blue}{\cite{github}}.

 \indent \textit{Clinical relevance}— Detection of Self-Stimulatory behaviors from recorded videos forms a key step towards the development of automated and cost-effective technology for screening, early diagnosis, and tracking of developmental disorders.
\end{abstract}

\section{INTRODUCTION} 
Autism Spectrum Disorder (ASD) is a neurological, developmental disorder characterized by the combination of social and cognitive impairments and repetitive sensory-motor actions, commonly referred to as self-stimulatory or stimming behaviors\cite{lord2018autism}.  
There is a wide variety and different forms of stimming actions - some examples include arm-flapping, headbanging, and spinning. These stimming actions help people with ASD to manage flooding sensory information and handle unsettling emotions by producing a calming effect in their bodies\cite{masiran2018stimming}.
Upon early detection of ASD, these behaviors can be mitigated while supporting self-regulation and remediating skill deficits which are critical to the overall development of the child \cite{tarbox2014handbook,oono2013parent}. 
The prevalence estimate of autism highlights that approximately one in 100 children are diagnosed with ASD worldwide~\cite{zeidan2022global}. The gold-standard technique to screen for ASD is through observational inventories. However, the major limitation of the observational approach is that some of the stimming behaviors might not be apparent during the assessment but may reach heightened levels at home. Also, parents' reports about the behavioral cues of the child may be very subjective.
In this regard, automated assessment of children's behavior from the recorded videos is an effective alternative for precise diagnosis and consistent tracking.

There have been multiple strides in the application of Machine Learning based methods to automate self-stimulatory behavior detection in children. The dataset that pioneered the use of videos for this task is the Self-Stimulatory Behaviors Dataset (SSBD) introduced by Rajagopalan et al. in 2013 \cite{rajagopalan2013self}.
 A time-distributed convolutional neural network (CNN) coupled with long short-term memory (LSTM) network was employed by Washington et al~\cite{10.1145/3411763.3451701} to perform a binary classification task of detecting headbanging within the videos of SSBD dataset. Similarly, Lakkapragada et al.\cite{2021arXiv210807917L} employed the hand landmarks detected by MediaPipe and feature representations from a MobileNetV2 model integrated into an LSTM layer to detect arm-flapping in a subset of videos from the SSBD dataset.  Min et al~\cite{min2017automatic} utilize multi-modal data comprising wearable sensors in conjunction with video data to accurately detect self-stimulatory behavior. \\

In this work, the authors augment the existing SSBD dataset~\cite{rajagopalan2013self} with a set of publicly available videos from YouTube which are annotated by a medical expert as detailed in section \ref{ssbdplus}. The updated dataset is employed to develop a novel, pipelined architecture, with the first stage dedicated to the detection and the second stage for the categorization of self-stimulatory actions viz. headbanging, spinning, and arm-flapping from the video snippets; as elaborated in section~\ref{pipeline}. 

This pipelined architecture has major advantages including better accuracy in identifying the self-stimulatory behaviors due to the large inter-class differences in the first stage and high amortized prediction speed; as the second stage of the model for behavior identification is recruited only if any valid self-stimulatory action is detected in the first phase.
This makes the pipeline easily deployable to mobile applications and offers more reliability in predictions. The \textit{no-class} category is introduced by the authors for the first time in the Self-Stimulatory Action Recognition scenario which has opened up the possibility for real-time and hands-free detection of such behaviors in the recorded videos in an uncontrolled environment. The pipelined deep learning architecture, and the associated dataset is made freely available for further research usage in~\cite{github}. Any researchers looking to use the dataset or models must first accept the licences present in the corresponding Github repositories. 

\thispagestyle{FirstPage}

\section{INTRODUCING SSBD+}
\label{ssbdplus}
The SSBD dataset, originally curated by Rajagopalan et al.\cite{rajagopalan2013self}, contains 25 videos collected from public domain websites like YouTube for each category of arm-flapping, headbanging, and spinning. The authors extend this dataset with 35 new videos, in the three aforementioned categories of stimming actions, gathered from YouTube by searching for the respective actions, for example with the prompt \textit{Headbanging autism actions in children}. These videos have an average duration of $\approx$90 seconds and are annotated by certified medical experts at the Bubbles Centre for Autism located in Bengaluru, India, and further stored in XML format as that of the SSBD dataset; hence being a natural extension to the same. The annotated dataset has been made open-source and hence the researchers are free to work with $\approx$45\% more data points in any future work for detecting self-stimulatory behaviors.

\section{Data Preprocessing}
\label{preproc}

The videos in the SSBD+ Dataset are sampled at $10$ fps and each frame is resized to the uniform dimensions of $100 \times 100$. From the sampled data, overlapping frames of size $40$ are grouped together to form video chunks of size $40 \times 3 \times 100 \times 100$ which acts as an input to the model. Each of the curated video chunks are then assigned with the class to which at least $75\%$ of the frames are belonging to. The audio streams are removed from each of the video chunks using the FFMPEG tool. In this work, $30\%$ of the video snippets of three stimming actions and no-class categories composed from the SSBD+ dataset are set aside for testing. The analysis of human body keypoints is a pivotal prerequisite and have enabled achieving the state-of-the-art~(SOTA) results in multiple tasks like Human Tracking, Gaming, Interpretation of Sign Languages and Human Action Recognition~\cite{munea2020progress}. Hence, a keypoint vector of size $40 \times 17$ is extracted from each of the chunks using the Movenet~\cite{tensorflow_2022} Lightning model and employed in the developed framework in order to boost the accuracy.

\section{Pipelined Architecture}
\label{pipeline}

\begin{figure}[thpb]
    \centering
    \includegraphics[scale=0.375]{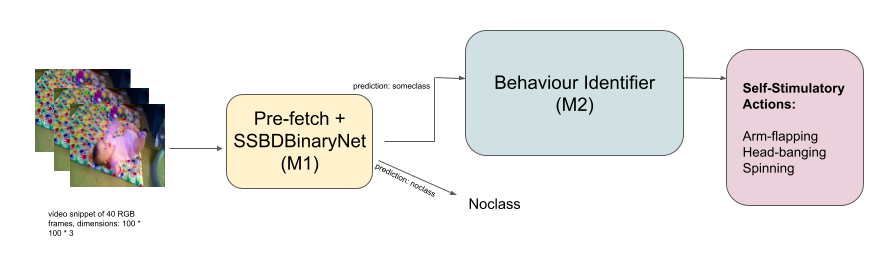}
    \caption{An overview of the SSBD Pipeline.}
    \label{pipelineoverview}
    \vspace{-4mm}
\end{figure} 
The dataset curated in the section \ref{ssbdplus} is observed to be highly unbalanced, with the ratio of about 7 \textit{no-class} video snippets for every video snippet showing any self-stimulatory action out of \textit{arm-flapping}, \textit{headbanging}, or \textit{spinning}. In order to avoid bias, a two-stage pipeline architecture is propounded as follows:

\begin{enumerate}
    \item \textbf{SSBDBinaryNet:} 

        Denoted as $M_1$, a binary video classification model detects the presence of any self-stimulatory action in the videos. The \textit{pre-fetch} model is also a part of $M_1$, and is described in \ref{m1desc}.
        
    \item \textbf{SSBDIdentifier:}
        A video classifier, denoted as $M_2$ identifies the action in the video snippets categorized as positive by $M_1$ for the presence of any stimming action. 
\end{enumerate}

The following subsections detail the architecture for each of the two models in the pipeline. 

\subsection{SSBDBinaryNet}
\label{m1desc}

As the primary concern of this work is to detect the stimming behavior in children, the authors of this paper filter the spatial portions of the video that corresponds to children in the frame. \\
In this regard, a \textit{pre-fetch} model comprising of the YOLOv7 object detector~\cite{wang2022yolov7} is employed to get the bounding boxes of all objects labeled as "person". Each spatial region of the image propounded by the Yolov7 for the presence of a human being is then classified as a Child or an Adult by a 
VGG19~\cite{vgg16} model finetuned on the Children vs Adults Classification dataset~\cite{kaggle_2022}. Although finetuning YoloV7 on this dataset was explored, the authors employed the former method of utilizing VGG19 as it showcased better accuracy and lower training time. 
For training the model, the Stochastic Gradient Descent~(SGD) optimizer with weight decay of $1E^{-5}$ is used with binary cross-entropy as the loss function. The LRFinder~\cite{silva_2020} tool is then used for estimating the optimal learning rate to aid in the fast convergence of the model on the training set. The final value of the learning rate used is $3.82E^{-02}$. The pre-fetch model is trained for 300 epochs with a batch size of 64, and achieved a test F1-score of $0.869$. For a particular frame of a video, the bounding box with the highest probability is chosen to be passed as input to the $M_1$ model. In case a particular frame of the video had no such bounding boxes, the crop region was chosen to be the largest bounding box measured as $height \cdot width$ found in the other frames.

The processed video snippets from the pre-fetch model have 40 frames each of three channels and dimension $100 \times 100$ which is then reshaped for convenience to $(40, 3, 100, 100)$ to form an input to the $M_1$ model. The $M_1$ architecture comprises of a (2 + 1)D convolution layer for feature extraction which was originally conceived by Tran et al~\cite{tran2018closer}. 
The (2 + 1)D convolution layer allows modularizing the task of extracting features from a video to the sub-tasks of spatial feature extraction by a two-dimensional convolutional layer and temporal feature extraction by a one-dimension convolutional layer. The (2 + 1)D convolution has relatively fewer computations and is less likely to overfit as compared to using a 3D convolution.

\begin{figure}[thpb]
    \centering
    \includegraphics[scale=0.17]{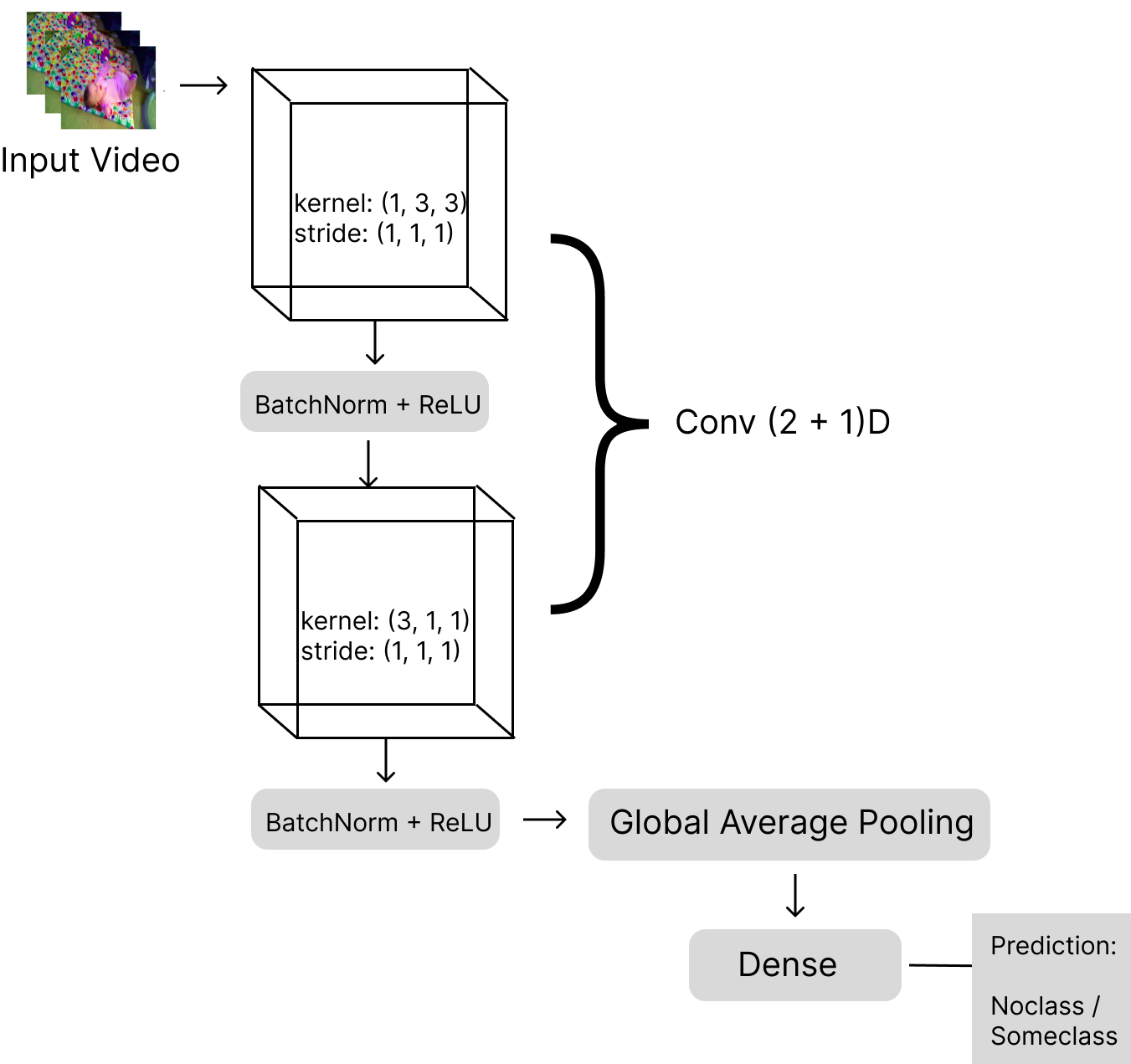}
    \caption{SSBDBinaryNet: Model Architecture}
    \label{m1}
    \vspace{-6mm}
\end{figure}

The feature extractor is followed by a 3D Batch Normalization layer and ReLU activation function in order to speed up the convergence. A global average pooling layer is stacked over 
the (2 + 1)D convolutional layer and the result is then passed through a fully connected, dense, feed-forward network to output the sigmoid probabilities for the binary class. 

If a video snippet is categorized into a self-stimulatory class by the $M_1$ model, then the original pre-processed video of shape $(40, 3, 100, 100)$ is fed as input to the $M_2$ model to identify the type of behavior exhibited in that snippet. And the snippets tagged negative for the stimming actions by the $M_1$ model are no longer processed and assigned with the label \textit{no-class}.

\subsection{SSBDIdentifier}
\label{m2desc}

\begin{figure}[thpb]
    \centering
    \includegraphics[scale=0.1]{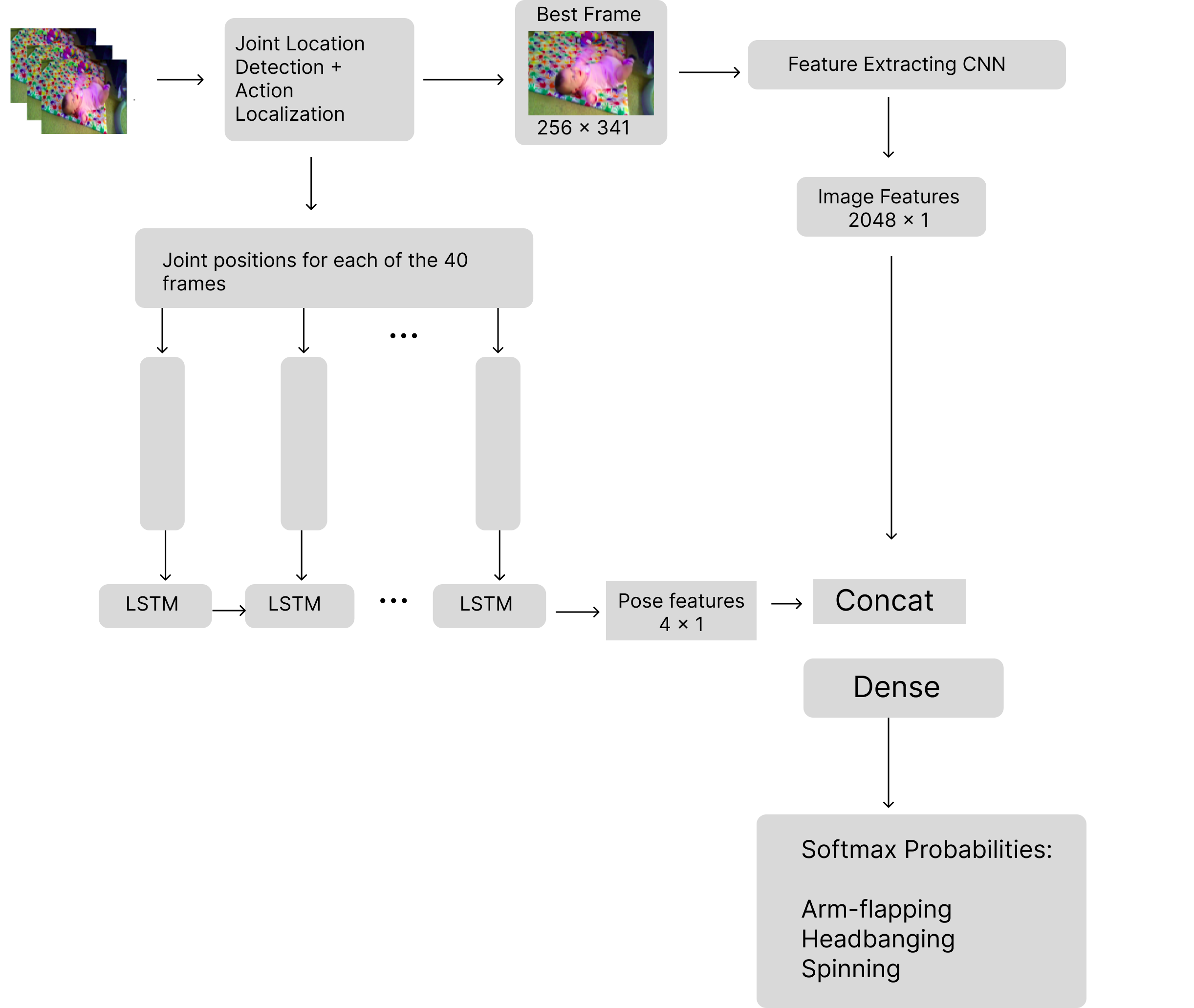}
    \caption{SSBDIdentifier: Model Architecture}
    \label{m2}
    \vspace{-4mm}
\end{figure}

\vspace{10pt}

Inspired by \cite{Lialin_2023}, the authors have exercised the concept of using a representative frame and the spatial location of key joints in all the frames for stimming behavior recognition for every video chunk. In this work, the \textit{representative} frame is selected as the one that has the maximum difference in joint locations from its previous frame in the video. The algorithm is shown in Algorithm \ref{frame-selection}. A complex feature map of shape $2048 \times 1$ is extracted from the penultimate layer of a pre-trained ResNet-18 \cite{he2015deep} for the \textit{representative} frame.

\begin{figure}
    \centering
    \includegraphics[scale=0.4]{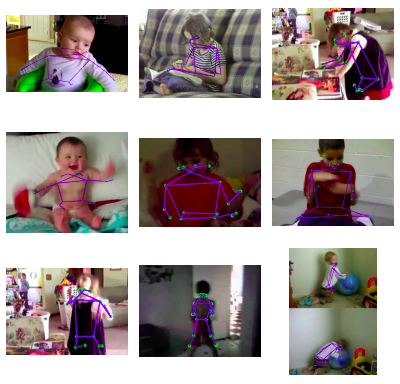}
    \caption{Pose points detected using MoveNet Thunder in a video of the SSBD set}
    \label{fig:pose_points}
\end{figure}

Then, the joint locations, from all the frames, vector of size $40 \times 17$, are then fed as an input in spatial order to the bi-directional LSTM layers to capture the temporal contextual features of shape $4\times 1$. The resultant vector is then concatenated to the spatial features of the representative frame, which gives it the shape of $2052 \times 1$.

The concatenation layer is followed by a fully-connected, feed-forward network that outputs the softmax probabilities for 3 classes namely \textit{arm-flapping}, \textit{headbanging}, and \textit{spinning}. It is also observed that the softmax probabilities are directly proportional to the intensity of the stimming behavior exhibited in the video and can be selected as a promising variable to track the outcomes of the therapy over time.  

 \begin{algorithm}
 \caption{Selecting the best frame for action recognition}
 \label{frame-selection}
 \begin{algorithmic}[1]
 \renewcommand{\algorithmicrequire}{\textbf{Input: }}
 \renewcommand{\algorithmicensure}{\textbf{Output:}}
 \REQUIRE Frames of the single video chunk $1$ to $40$ in the playing order (F)
 \REQUIRE Joint coordinates (J) detected in each frame of the chosen video chunk $1$ to $40$ (in the same order as in F) 
 \ENSURE  Index of the best frame to be evaluated by the model
 \\ \textit{Initialisation} :
  \STATE $maxDiff = 0$
  \STATE $maxFrameIdx = -1$
    \FOR {$t = 1$ to $t = 39$}
        \STATE $diff = ||J[t] - J[t + 1]||$
        \IF {$maxDiff < diff$}
                \STATE $maxDiff = diff$
                \STATE $maxFrameIdx = t$  
    \ENDIF
    \ENDFOR
 \RETURN $F[maxFrameIdx]$ 
 \end{algorithmic} 
 \end{algorithm}

\begin{figure}
    \centering
    \includegraphics[scale=0.4]{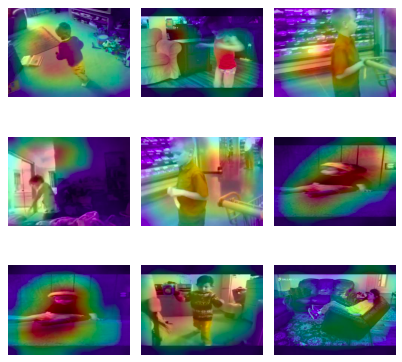}
    \caption{GradCAM~\cite{Selvaraju_2019} images of children from the SSBD+ set using XceptionNet~\cite{chollet2017xception}. The magnitude of the activation is the highest near the child's face and body, showing the higher importance given to the area of the frame by the feature extractor}
    \label{fig:gradcam_image}
    \vspace{-6mm}
\end{figure}

\section{Model Training and Results}
\label{results}


The $M_1$ model is trained for 240 epochs with a batch size of 128. For training the model, the Stochastic Gradient Descent~(SGD) optimizer having a momentum of 0.3 and weight decay of $1E^{-5}$ is used with binary cross-entropy as the loss function. The LRFinder \cite{silva_2020} is then used for estimating the optimal learning rate 
and the final value of the learning rate was chosen to be $2.31E^{-03}$. The $M_2$ model is trained with a batch size of 64 for 300 epochs. An SGD optimizer having a weight decay of $8.29E^{-5}$ is used with categorical cross-entropy as the loss function for training the $M_2$ model. The optimal learning rate for the $M_2$ model is estimated to be $8.29E^{-01}$ by LRFinder. Both of the models are then evaluated with a test set encompassing the curated video snippets belonging to all the four possible output categories namely \emph{no-class}, \emph{arm-flapping}, \emph{headbanging}, and \emph{spinning}. Table~\ref{metrics} summarises the accuracy and F1 score for the two models employed in the pipeline. 
An overall accuracy of around 81\% is achieved from the proposed pipeline model that is targeted for real-time and hands-free automated diagnosis, which does not exist in the current SOTA scheme of methods.

\begin{table}[!h]
\vspace{-2mm}
\caption{Accuracy and F1-score of the pipelined models over the newly curated set from YouTube.}
\vspace{-4mm}
\label{metrics}
\begin{center}
\begin{tabular}{|c|c|c|}
\hline
\textbf{Model} & \textbf{F1-Score} & \textbf{Accuracy} \\
\hline
SSBDBinaryNet & 0.819 & 0.811 \\
\hline
SSBDIdentifier & 0.789 & 0.812 \\
\hline
\end{tabular}
\end{center}
\vspace{-6mm}
\end{table}

\begin{table}[!h]
\vspace{-2mm}
\caption{Model footprints of the models in the pipeline}
\vspace{-4mm}
\label{footprints}
\begin{center}
\begin{tabular}{|c|c|c|}
\hline
\textbf{Model} & \textbf{Total \#Weights} & \textbf{Learnable \#Weights} \\
\hline
PrefetchVGG & 143,951,722 (143.9M) & 273,474 (273.4K) \\
\hline
{SSBDBinaryNet CNN} & 38,265 (38.2K) & 38,265 (38.2K) \\
\hline
SSBDIdentifier & 20,861,480 (20.8M) & 6,783 (6.7K) \\
\hline
\end{tabular}
\end{center}
\vspace{-6mm}
\end{table}

\section{Boosting Robustness of the Pipeline}
In \ref{pipeline}, the authors have proposed a pipelined architecture that first detects if the child performs one of the three actions: arm-flapping, headbanging, and spinning. If such an action is present it then identifies the specific action being performed. As indicated in Table \ref{metrics}, the detection model filters videos containing one of the three actions with an accuracy of 81\%. The incorrect predictions can be minimized by considering two cases: false negatives and false positives.  

In the case of false negatives, the authors propose to analyze $k$ contiguous video snippets at a time as against independently passing the snippets through the pipeline. Since we analyze the video by snippets, we run $M_1$ on all these snippets. In order to reduce the error propagation in the pipeline due to misclassification by this model, the authors propose considering $k = 2$ consecutive snippets' outputs, and if either of those outputs detects the presence of a self-stimulatory behavior, we pass both the snippets to the next ($M_2$) model in the pipeline. Since each action is performed in a certain contiguous period of time in a video, one can take consecutive snippets to find the presence of the action instead of evaluating each snippet independently.  

In the case of false positives, where snippets with none of these actions are passed to $M_2$, the authors propose to threshold the softmax probabilities predicted by $M_2$. That is, if the probability of all of the classes is less than $0.33 + \delta$, then it is earmarked as \emph{noclass}. Pruning the threshold values to find the exact value of $\delta$ ($\delta \in R^{+}$), constitutes the future tasks of this work. 

The methodology of processing \textit{windows} of chunks (contiguous segments) has been detailed in Algorithm \ref{windowing_algo}. In the event of the number of chunks $C$ being divisible by the window size chosen $k$, this algorithm goes through every chunk's output efficiently.

 \begin{algorithm}
 \caption{Processing $k$ contiguous video chunks in $M_1$ - \textit{windowing}}  \label{windowing_algo}
 \begin{algorithmic}[1]
 \renewcommand{\algorithmicrequire}{\textbf{Input: }}
 \renewcommand{\algorithmicensure}{\textbf{Output:}}
 \REQUIRE Video chunks indexed from $1$ to $C$
 \REQUIRE Window size $W$
 \ENSURE  List of chunk indices to be passed to $M_2$
 \\ \textit{Initialisation} :
  \STATE $chunkIndex = 1$
  \STATE $O = \Phi$
  \WHILE {$chunkIndex \leq C - W + 1$}
    \FOR {$t = chunkIndex$ to $t = chunkIndex + W$}
        \STATE $output = M_1$[40 frames in chunk $t$]
        \IF {$output == $"Action"}
                \STATE $O = O \bigcup \{chunkIndex, \dots, chunkIndex + W\}$
                \STATE GOTO 11           
    \ENDIF
    \ENDFOR
  \STATE $chunkIndex = chunkIndex + W$
  \ENDWHILE
 \RETURN $O$ 
 \end{algorithmic} 
 \end{algorithm}

\section{Discussions}
\subsection{Alternate approaches}
There were several approaches tried for the detection and identification task. One of the initial approaches involved fine-tuning the state-of-the-art(SOTA) transformer for action recognition, DeVTr (Data Efficient Video Transformer for Violence Detection) \cite{abdali2021data} with the SSBD dataset. However, the fine-tuned model failed to generalize to the unseen test videos taken from the newly curated YouTube set as well as the data from SSBD if the child under observation in the test video is different from that of the train videos. 
Further, various feature extractors such as C3D \cite{tran2015learning} and MXNet \cite{wang2016temporal} were also used which gave subpar results. 

\subsection{Scope for Post-processing}

If this architecture is used on recorded videos where the requirements for latency are laxer, the effective results of the pipeline can be improved through \textit{post-processing} methods. The authors recommend analyzing the predictions of the consecutive video snippets to rectify any incorrect detections and misclassifications of $M_1$ and $M_2$.

\subsection{Pre-fetch model: Design Choices}
Although finetuning YoloV7 for this task was explored, the authors employed the method of utilizing VGG19 as it showcased better accuracy and lower training time. The authors also experimented with different YoloV7 variants and decided on \textit{YoloV7-X} as it provided the best observed FPS among the variants tried, as detailed in Table \ref{yolo_fps}. These figures are averaged over 75 video chunks selected at random, and were observed on the Tesla P100-PCIE-16GB GPU.

\begin{table}[!h]
\vspace{-2mm}
\caption{Observed Frames per second (FPS) of Yolov7 variants}  \label{yolo_fps}
\vspace{-4mm}
\label{speed}
\begin{center}
\begin{tabular}{|c|c|}
\hline
\textbf{Model Variant} & \textbf{Average FPS} \\
\hline
Yolov7-X & 77.75 \\
\hline
Yolov7-E6 & 71.49 \\
\hline
Yolov7-D6 & 65.82 \\
\hline
\end{tabular}
\end{center}
\vspace{-6mm}
\end{table}

\subsection{Towards low-latency stimming behaviour detection models}
The pipeline architecture of the model has had a significant contribution to the low inference latency of the model, due to the selective processing of videos to identify the SSBD action first before classifying the action itself, thereby reducing the amortised inference time. 

In the event of lower compute resource requirements in a setting that warrants the use of $M_1$, the authors recommend omitting the Yolov7 and Pre-fetch model and input the frames directly to $M_1$. Using just the $M_1$ model without the Yolov7-VGG backbone enables $1.8x$ faster inference on average, researchers can note the component-wise FPS of $M_1$ in Table \ref{prefetch_speed}. However, the F1 score of this smaller system ($0.740$) is lower than that of the SSBDBinaryNet ($0.819$). These figures are same chunks used to generate table \ref{yolo_fps}.

\begin{table}[!h]
\vspace{-2mm}
\caption{Observed Frames per second (FPS) of the components of $M_1$ and Pre-fetch}
\vspace{-4mm}
\label{prefetch_speed}
\begin{center}
\begin{tabular}{|c|c|}
\hline
\textbf{Component} & \textbf{Average FPS} \\
\hline
Yolov7-X & 77.75 \\
\hline
Pre-fetch VGG network & 110.84 \\
\hline
$M_1$ CNN & 235.31 \\
\hline
\end{tabular}
\end{center}
\vspace{-6mm}
\end{table} 

\vspace{5pt}

As demonstrated by Chen et al. in ~\cite{ASD_mobility}, \textit{distillation} can be a powerful technique in the field of Autism Spectum Disorder Screening. In this regard, the authors present their experimental setting for distilling the "learnings" of a \textit{teacher model} into a smaller \textit{student model}. \\
The \textbf{teacher model} consists of a pre-trained Resnet-18 backbone followed by a Bi-LSTM block and a MultiHead Attention block. The resultant features are then passed through three fully-connected (FC) layers. The Resnet backbone's final layer was trainable and the other layers were frozen. In total, this model has $23,836,579$ (23.8M) learnable weights. \\
The \textbf{student model} consists of a pre-trained Resnet-18 backbone followed by a LSTM block and two fully-connected layers. To reduce the number of learnable parameters, the entire Resnet backbone is frozen and the number of LSTM cells is halved from the teacher model. In total, this model has $8,911,107$ (8.9M) learnable weights. \\
Both models do not take Movenet features as inputs and are trained to identify the action in the video snippets classified as positive by $M_1$. The loss function for the student model is a weighted sum of the cross-entropy loss $L_{CE}$ of its own logits and the ground truth labels; and the soft target loss $L_{SOFT}$ (between the logits of the student model and the teacher model) as described in ~\cite{hinton2015distilling}. The temperature value augmented to the softmax outputs, $T$ = $2$. \\ 
The overall loss function chosen is $L_{DISTILLATION} = $ $0.25 \cdot L_{SOFT} + 0.75 \cdot L_{CE}$. \\ 
The student model was trained with this extra objective while the teacher model's weights were frozen. Despite having only $37.38\%$ learnable weights, the student model was able to reach $80.89\%$ of the test F1-score of the teacher model. 

The authors note that this experiment shows bright potential for model distillation in the domain of self-stimulatory behaviour detection and future work includes developing novel distilled models suitable for low-latency deployment scenarios.

\subsection{Ablation Study for $M_2$}

Currently, the $M_2$ model in the pipeline uses a $representative$ frame from the given video and Movenet pose coordinates to identify the self-stimulatory actions. However, multiple explorations were carried out before concluding to use only a single frame representing an entire video. Table \ref{ablation_study} shows the performance in each ablation case. The following ablations were carried out by the authors:\\ \\
\textbf{Ablation 1:} Using all 40 frames of a video chunk \\ In this method, for each video chunk, the authors passed all frames of the video chunk in the form of a matrix of dimensions (40, 256, 256) along with the pose coordinates of dimensions (40, 34). Each frame was passed through the convolutional feature extractor, to get the matrix of size (40, 512). The authors then concatenated the spatial features along with Movenet features, to get the matrix of dimensions (40, 546), which is passed through the bi-directional LSTM to extract the temporal features.  \\  \\
\textbf{Ablation 2:} Using a single representative frame for each video chunk\\ This is the approach authors recommend in Section \ref{m2desc}. The model shows an improvement of $0.137$ in the F1-score when this method is used as compared to \textit{Ablation 1}.

\begin{table}[!h]
\vspace{-2mm}
\caption{Performance of each ablation carried out by the authors in $M_2$}
\vspace{-4mm}
\label{ablation_study}
\begin{center}
\begin{tabular}{|c|c|}
\hline
\textbf{Ablation} & \textbf{F1-Score} \\
\hline
All Frames & 0.652 \\
\hline
Single Representative Frame & 0.789 \\
\hline
\end{tabular}
\end{center}
\vspace{-6mm}
\end{table} 

\section{Conclusion}
In the proposed work, a novel deep learning-based pipelined architecture to automatically screen self-stimulatory behaviors from raw videos is introduced. In the earlier reported works on stimming behavior categorization, the video segments containing any self-stimulatory behaviors were manually cropped and fed as input to the classification model which is not suitable for real-time analysis of videos. And to the best of the authors knowledge, this is the first time a no-class category is introduced and this enables real-time and completely autonomous detection of stimming behaviors. 
The authors of this paper also explore alternative schemes to use the pipeline in the case of deployment into a low-latency requirement-driven environment and attain an accuracy of 81\%. 

Additionally, new videos have been provided as an addition to the SSBD dataset. This has resulted in a $\approx$45\% increase in the number of data points to researchers. The format of the dataset has been made similar to that of SSBD to promote ease of use by the community.
The proposition of using the softmax outputs as confidence scores to track action intensities over time is an interesting future work for this task.
All of the source code, data, and other relevant material is made freely available in~\textcolor{blue}{\cite{github}}.

\section{Acknowledgement}
The authors thank the psychiatrists at Bubbles Center for Autism, India for providing us with annotations for the videos in the SSBD+ dataset.
The authors acknowledge the support and the research grant from IBM GUP.

\balance
\bibliographystyle{IEEEtran}

\bibliography{references}

\end{document}